\documentclass[review]{elsarticle}
\usepackage{lineno,hyperref}
\usepackage{lscape}
 \usepackage{ulem}
 \usepackage{color}

\journal{Pattern Recognition}

\begin{document}
	
	\begin{frontmatter}
		
		\title{Three-Stream Fusion Network for \\First-Person Interaction Recognition}
		
		
		\author[mymainaddress]{Ye-Ji Kim}
		\author[mymainaddress]{Dong-Gyu Lee}

		\author[mysecondaryaddress]{Seong-Whan Lee\corref{mycorrespondingauthor}}
		\cortext[mycorrespondingauthor]{Corresponding author}
		\ead{sw.lee@korea.ac.kr}
		
		\address[mymainaddress]{Department of Computer and Radio Communications Engineering, Korea University, Anam-dong, Seongbuk-gu, Seoul 02841, Korea}
		\address[mysecondaryaddress]{Department of Artificial Intelligence, Korea University, Anam-dong, Seongbuk-gu, Seoul 02841, Korea}
		
		\begin{abstract}
			First-person interaction recognition is a challenging task because of unstable video conditions resulting from the camera wearer’s movement. For human interaction recognition from a first-person viewpoint, this paper proposes a three-stream fusion network with two main parts: three-stream architecture and three-stream correlation fusion. The three-stream architecture captures the characteristics of the target appearance, target motion, and camera ego-motion. Meanwhile the three-stream correlation fusion combines the feature map of each of the three streams to consider the correlations among the target appearance, target motion, and camera ego-motion. The fused feature vector is robust to the camera movement and compensates for the noise of the camera ego-motion. Short-term intervals are modeled using the fused feature vector, and a long short-term memory(LSTM) model considers the temporal dynamics of the video. We evaluated the proposed method on two public benchmark datasets to validate the effectiveness of our approach. The experimental results show that the proposed fusion method successfully generated a discriminative feature vector, and our network outperformed all competing activity recognition methods in first-person videos where considerable camera ego-motion occurs.
		\end{abstract}
		
		\begin{keyword}
			First-person vision \sep first-person interaction recognition \sep three-stream fusion network\sep three-stream correlation fusion \sep camera ego-motion.
		\end{keyword}
		
	\end{frontmatter}

	\section{Introduction}
	
	Despite the increasing research on computer vision, the task of understanding human activity in videos remains a challenging task. Recent approaches based on deep learning techniques have achieved significant progress in third-person activity recognition \cite{VECTOR2017,SPATIO-RESNET, FUSION,TWO, PYRAMID}. In third-person videos, the camera is fixed and at a large distance from people and objects. Many researchers have also studied human activity recognition in first-person video \cite{sport,LETTER, CVPR2016,RYOO2013, SUB-EVENT}. First-person video is captured by a camera mounted on a person or object. The video has a unique characteristics called camera ego-motion, which is not usually seen in third-person video. When the video includes camera ego-motion, the appearance of the target subjects is easily distorted and the motion vectors are disordered. Therefore, the recognition of activities in first-person video requires an appropriate approach tailored to these particular characteristics. In this paper, we analyze first-person video frames and focus especially on the interaction between a camera wearer and a human subject. 
	
	\begin{figure}[t]
		\begin{center}
			\includegraphics[width=0.8\linewidth]{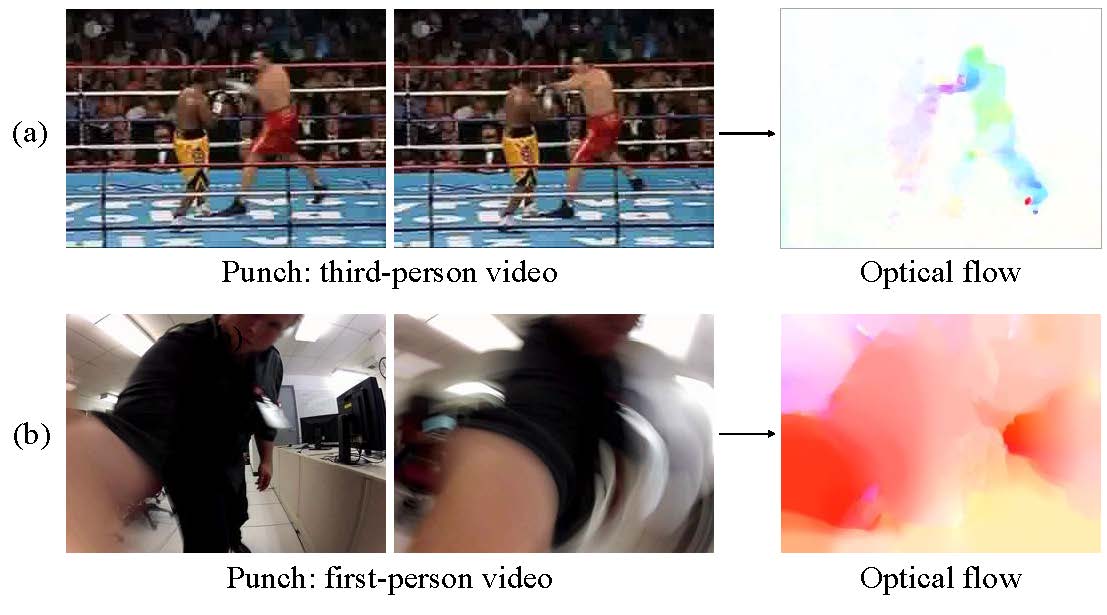}
			\caption{\label{optical}Punch action in third-person and first-person videos. The two video clips have different characteristics in terms of appearance and motion. (a) The optical flow is extracted from a third-person video, where the camera is fixed. (b) The optical flow is extracted from a first-person video where the camera shakes considerably.}
		\end{center}
	\end{figure}
	
	In Figure 1, the optical flow extracted from two consecutive RGB frames shows various motion vectors in each third-person and first-person video. As shown in Figure 1(a), motion vectors in the third-person video appear around the region where people are positioned because the camera is fixed and stationary. However, Figure 1(b) shows complex motion vectors for the target and the camera wearer. The motion vectors appear close to the region where the target is positioned and the entire region of optical flow. Camera ego-motion renders analyzing the appearance and motion characteristics of the target in first-person video more challenging for three reasons: (1) it is difficult to build discriminative motion features from the complex motion vector in first-person video; (2) target appearance, which is an important feature for analyzing the target’s behavior, is severely distorted because of the camera wearer’s movement; (3) camera ego-motion can occur when the camera wearer moves, regardless of the target’s actions, and its data can be noisy when first-person interactions are analyzed. 
	
	This paper proposes a three-stream fusion network to recognize interactions in first-person video where large amounts of camera ego-motion occur. The proposed method is composed of two main parts: three-stream architecture and three-stream correlation fusion (TSCF). The three-stream architecture consists of target appearance stream, target motion stream, and ego-motion stream. The target appearance stream and the motion stream capture the appearance features and motion features of the target, respectively, and the ego-motion stream captures features of the camera wearer's movement. The camera ego-motion is an important clue to the camera wearer’s movement because the wearer’s pose in first-person video is unknown. To generate robust features for the camera ego-motion, the TSCF considers two types of correlations. It considers the correlation between the target appearance and motion to utilize the spatiotemporal relationship. This relationship complements the target’s appearance and motion features, which are distorted by the camera wearer's movement. To consider the problem of the camera wearer's movement being noisy, the TSCF also uses the correlation between the target and camera ego-motion to determine whether the camera wearer's movement is caused by the target's action. In other words, we can determine whether the camera ego-motion is an important clue for analyzing first-person interactions between the target and the camera wearer. 
	
	The main contributions of this paper are summarized as follows: (1) we propose a novel deep learning framework called the three-stream fusion network. The proposed network is specialized in extracting discriminative features and considers camera ego-motion an important feature for analyzing the camera wearer’s movement; (2) we also introduce a fusion method called TSCF, which considers the correlations between the target's appearance, motion and the camera ego-motion. The proposed fusion method creates robust features mitigate the effects of the camera ego-motion; (3)wWe show that our proposed method outperforms state-of-the-art activity recognition methods using the JPL First-Person Interaction dataset and the UTKinect-FirstPerson dataset.
	
	\section{Related Works}
	
	There has been a great deal of progress in human activity recognition in video captured from a third-person viewpoint. Early work contributed hand-craft features to feature representation for activity recognition \cite{OPTICAL, park2004qualitative, HOF, IDT, roh2010view, suk2008recognizing, roh2000multiple, suk2011network}. Some studies suggested various methods, such as support vector machine (SVM) \cite{laptev2004recognizing, xi2002facial}, unsupervised learning \cite{du2016stacked}, and multi-label learning \cite{du2017robust} to improve recognition performance. In more recent research, a significant performance increment was achieved using deep ConvNet \cite{VECTOR2017, SPATIO-RESNET, FUSION, AAAI2018, xu2018spectral}. Simonyan \textit{et al.} \cite{TWO} proposed a two-stream architecture composed of a spatial and a temporal stream to capture appearance and motion features separately. Feichtenhofer \textit{et al.} \cite{FUSION} proposed a number of fusion methods to combine the convolutional neural networks (CNNs) of the two-stream architecture. The spatial and temporal streams were combined at the convolutional layer of the CNNs to take advantage of spatiotemporal features. Other recent research studies on activity recognition utilized long short-term memory (LSTM) to handle the time-series information of video \cite{LRCN, RPAN, RNN2017}. To overcome the viewpoint dependency, Roh \textit{et al.} \cite{roh2010view} proposed a volume motion template (VMT) and a projected motion template (PMT). However, these templates are not optimized for first-person video, which has different characteristics, such as camera ego-motion, multiple visual scales, and proceeding events (\textit{e.g.} after the target delivers a punch, the camera wearer falls down).
	
	
	
	

	The research on activity recognition in first-person video can be divided into three categories. (1) The action recognition of a camera wearer  \cite{LETTER, DOG, BUDGET}. This research is focused on actions related to ``What am I doing alone?'' such as \textit{walking}, \textit{running}, \textit{standing}, and \textit{going up the stairs}. Ryoo \textit{et al.} \cite{RYOO2015} proposed a novel feature representation to capture camera ego-motion and continuously track detailed changes while noisy data are suppressed. Abebe \textit{et al.} \cite{abebe} used stacked spectrograms obtained from mean grid-optical flow vectors and the displacement vectors of the intensity centroid to represent various motions in video. (2) The interaction recognition between a camera wearer’s hand and objects \cite{ECCV2018, DESCRIPT2016,PR, TIP,CASCADE2016}. This research is focused on actions related to ``How do I interact with what type of objects?'' such as \textit{take jam}, \textit{close freezer}, and \textit{cut mushroom}. Ma \textit{et al.} \cite{CVPR2016} proposed an egocentric activity deep network using ObjectNet and ActionNet, where ObjectNet captures the object of an interesting region and the ActionNet captures the motion of both the camera wearer and the handled objects. Li \textit{et al.} \cite{DELVING2015} proposed a set of egocentric features, such as hands, gaze, object features, and head motion, and showed the manner in which they can be combined. (3) The interaction recognition between a camera wearer and a human \cite{RYOO2013, SUB-EVENT, KERNEL,ROBOT,IJCV2016,ICCVW2017, MICRO}. This research is focused on action related to ``What is he doing to me?'' such as \textit{hand shake}, \textit{pet}, \textit{punch}, \textit{hug}, and \textit{throw} actions. In this paper, we especially focus on the interaction between a camera wearer and a human. Ryoo \textit{et al.} \cite{RYOO2013} used hand-crafted features in first-person video, such as optical flow, to capture global and local motion descriptors. Early recognition methods \cite{ROBOT} inferred an ongoing activity at the early stages in first-person video obtained from a robot’s viewpoint. Histograms of time-series gradients were used to consider the event history in a video. Zaki \textit{et al.} \cite{SUB-EVENT} modeled sub-event dynamics to track their relations over time. After the local temporal structure of sub-events was encoded, the global temporal structure of the video was encoded. Sudhakaran \textit{et al.} \cite{ICCVW2017} used a pair of CNNs to capture the frame-wise features of two consecutive input images and convolutional LSTM to aggregate them.

	Our proposed network differ from previous first-person interaction recognition methods in that, in addition to the appearance and the motion of the target, it uses the camera ego-motion a feature type. These three feature types are combined using our fusion method. The method considers the correlations between the three types of features and thus it is rendered more robust to camera ego-motion. We show that this is an important design choice for analyzing first-person interaction, where a large amount of movement is generated by the camera wearer.

	\section{Three-Stream Fusion Network} 
	An overview of our three-stream fusion network is shown in Figure 2. Our proposed network consists of the three-stream architecture, TSCF, and an LSTM model. Our three-stream architecture captures features for the target’s appearance, motion, and the camera ego-motion. The output feature maps from the three streams are combined by the proposed TSCF. The LSTM model classifies the video considering the temporal dynamics of TSCF features.

	\begin{figure*}[t]
		\begin{center}
			\includegraphics[width=1.0\linewidth]{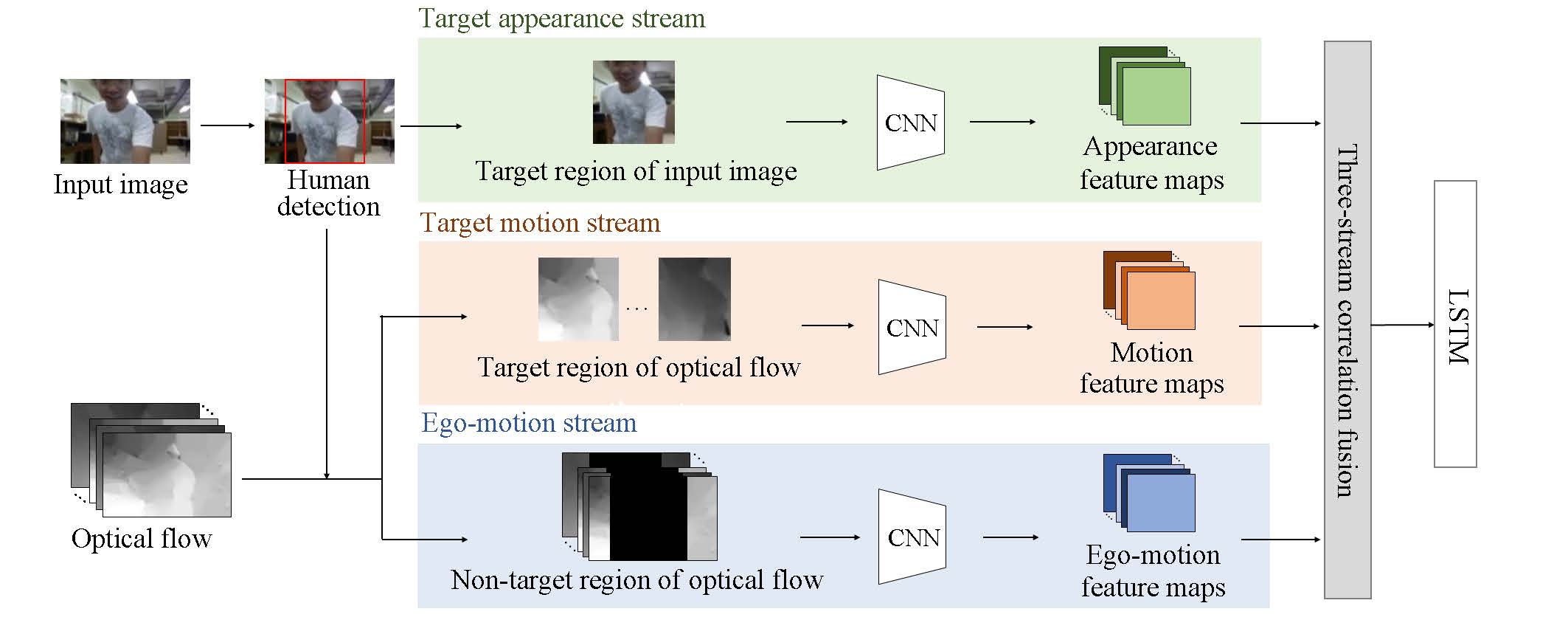} 
			\caption{\label{framework}Overview of the three-stream fusion network. Our proposed network is composed of three-stream architecture, the three-stream correlation fusion (TSCF), and a long short-term memory model. Each stream of the three-stream architecture respectively extracts appearance, motion, and ego-motion feature maps. Then, the proposed TSCF combines the output feature maps of the three streams. The LSTM model takes the fused features as an input value to classify the video class.}
		\end{center}
	\end{figure*}

	\subsection{Three-Stream Architecture}
	The proposed three-stream architecture is composed of the target appearance stream, the target motion stream, and the ego-motion stream, which respectively consider each of the characteristics individually. Intuitively, the target appearance stream and the target motion stream capture the appearance and motion cues of the target, and the ego-motion stream captures the cues of the camera ego-motion.

	The target appearance stream uses the target region of an input image. To focus on the appearance of the target, we exclude as much of the background region of the input image as possible. First, an input image captured at time $t$ is taken to detect the target using the Faster R-CNN \cite{rcnn}, which can capture the coordinates of the human region. Then, we crop the region where the target (human) is located from the input image using the coordinates of the human region. The cropped region is entered into the target appearance stream. The target motion and ego-motion streams are designed under the consideration that the movement of the camera wearer generates motion vectors in the background region of the optical flow. Thus, these two streams use different regions of the optical flow to distinguish the target motion and the camera ego-motion in video. First, we stack the optical flows extracted from consecutive video frames. The stack consists of 20 optical flows of two directions from time $t$ to $t+9$. Next, we divide each optical flow with the target region and background region(non-target region). In this case, the coordinates of the human region detected from the video frames are utilized to divide the optical flow. The target motion stream uses the target region of the optical flow as an input value. The aim is exclude the vectors of the camera ego-motion as much as possible from the optical flow and focus on the target motion. To create an input value of the ego-motion stream using the non-target region of the optical flow, we fill the zero values at the target region of the optical flow to focus on the camera ego-motion.

	When considerable camera ego-motion occurs and the target appearance is seriously distorted, the Faster R-CNN occasionally fails to detect the human region in video frames. In this case, we set the zero matrix size of $100\times100$ as an input value of the target appearance stream and the target motion stream. In addition, the ego-motion stream uses the optical flow itself as an input value.
	
	The CNN of each of the three streams generates deep feature maps of the target appearance, target motion, and camera ego-motion. Then, each of the final convolutional layers of the CNNs is extracted as a feature map. As a result, the target appearance stream generates the appearance feature maps $f^{app}$ with the target region of the input image, the target motion stream generates the motion feature maps $f^{mot}$ with the target region of the optical flow, and the ego-motion stream generates the ego-motion feature maps $f^{ego}$ with the non-target region of the optical flow. We denote the feature maps by $f\in R^{512\times7\times7}$. 
	


	\subsection{Three-Stream Correlation Fusion}
	We propose a fusion method called TSCF to create robust features to mitigate effects of the camera ego-motion. Figure 3 shows an overview of TSCF. The appearance, motion, and ego-motion feature maps are generated from the three-stream architecture. First, we sum the appearance and motion feature maps to consider the correlation between the target appearance and the target motion. Then attentional pooling is applied to the appearance feature maps with the sum of the appearance and motion feature maps. This is aimed to complement the target appearance, which is distorted by the camera ego-motion. After the maximum values are taken from the combined attention, we generate a channel-wise max pooling vector, which is intended to create a discriminative feature even if the camera is shaken.

			\begin{figure}[h]
		\begin{center}
			\includegraphics[width=0.8\linewidth]{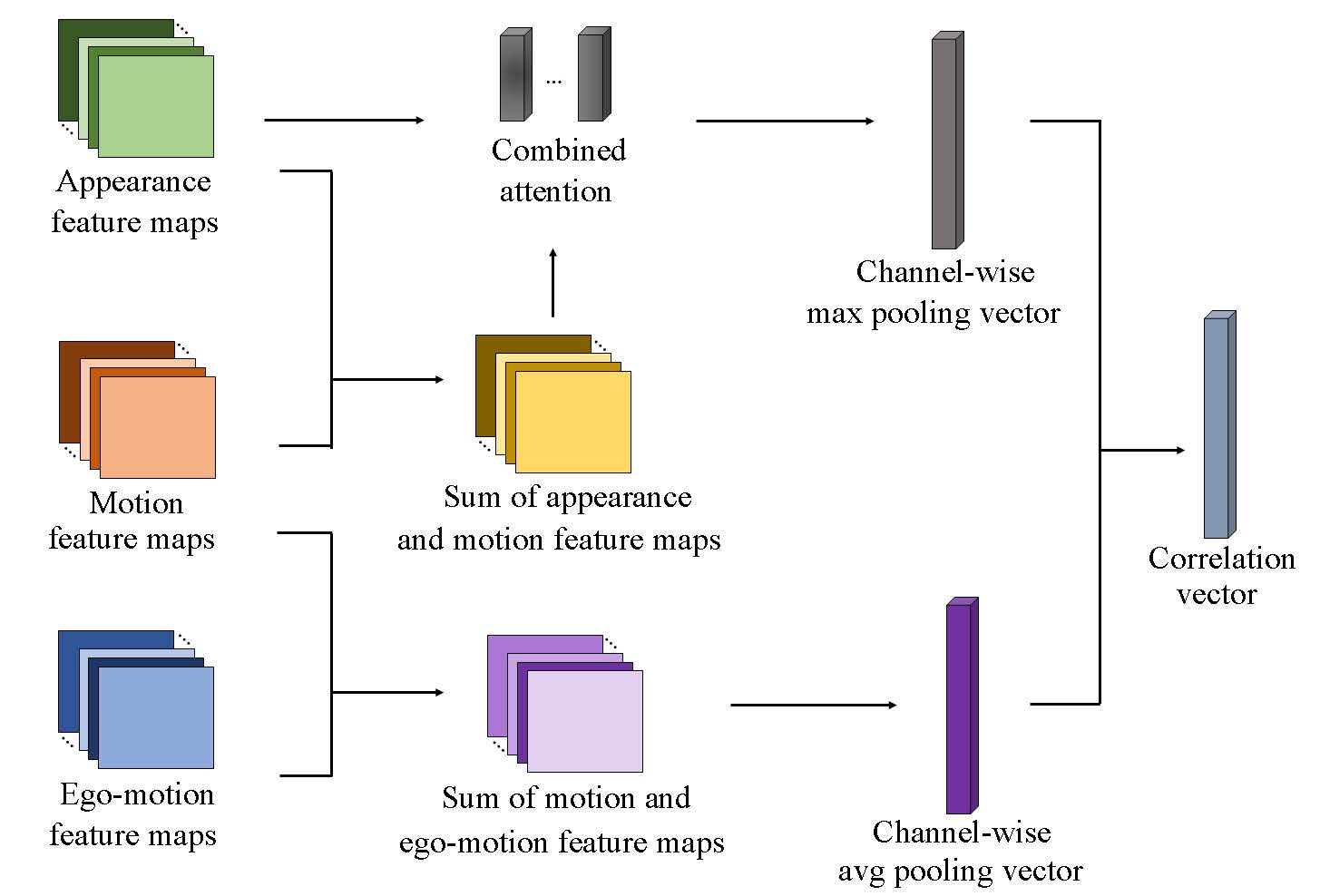} 
			\caption{\label{framework}Overview of the three-stream correlation fusion (TSCF). TSCF fuses the output feature maps of the three-stream architecture, considering the correlations between the target appearance, target motion, and camera ego-motion.}
		\end{center}
	\end{figure}
		
		Second, the sum of the motion and ego-motion feature maps is calculated to consider the correlation of the motion vectors resulting from the interaction of the target human and the camera wearer. By considering this correlation, we can determine whether the camera wearer's movement is caused by the target. We take the average values from the sum of the motion and ego-motion feature maps and create the channel-wise average pooling vector, because the sum of the motion and ego-motion feature maps include a noise value when the camera shakes considerably. We also consider the correlation between the target's action and the camera wearer's movement. A correlation vector is generated from the sum of the channel-wise max pooling vector and the channel-wise average pooling vector. It considers the correlation between the target's action and the camera wearer's movement.

	To fuse the three types of feature maps, we use the sum fusion \cite{FUSION}, which can combine two different kinds of feature maps $f^x$, $f^y$ at the same channel $d$ and spatial location $i$,$j$:
	\begin{eqnarray}
	f^{sum}_{d,i,j} = f^x_{d,i,j}+f^y_{d,i,j}
	\end{eqnarray}
	where $1\leq d\leq 512$, $1\leq i\leq 7$, $1\leq j\leq 7$. $x$ and $y$ refer to specific feature maps. First, the appearance and motion feature maps are combined to make $f^{spa}$, which contains the spatiotemporal cues of the target's action. Then, the sum of the motion and ego-motion feature maps is calculated in the same manner to produce $f^{tmo}$.
	

	We also calculate the attentional pooling \cite{ATTENTION} to compensate for the target appearance and target motion, which are distorted by the camera ego-motion. Because the attentional pooling method can render feature maps more related to certain tasks, we utilize it to make the sum of the appearance and motion feature maps more robust to the camera ego-motion. The feature map $f^{spa}_d$ is taken from the $f^{spa}$ at channel $d$, and the matrix multiplication of $f^{spa}_d$ is performed as follows:
	\begin{eqnarray}
	X = {f^{spa}_d}{f^{spa}_d}^{T}
	\end{eqnarray}
	where $X\in R^{7\times7}$. Then, we calculate a matrix $W=\mathbf{ab}^{T}$ by performing the rank-1 approximation from $X$. In this case, the singular value decomposition (SVD) function in the numpy library \cite{svd} was used to perform the $n$-rank approximation. We perform the matrix multiplication with the $\mathbf{a}$ of $W$ and the appearance feature map at channel $d$ to make a top-down attention $f^{app}_d\mathbf{a}$. In addition, the $\mathbf{b}$ of $W$ is used to perform the matrix multiplication with the appearance feature map to make a bottom-up attention $f^{app}_d\mathbf{b}$. A combined attention $c^{att}_d$ is generated by performing element-wise multiplication on $f^{app}_d\mathbf{a}$ and $f^{app}_d\mathbf{b}$:
	\begin{eqnarray}
	c^{att}_d= f^{app}_d\mathbf{a}\circ f^{app}_d\mathbf{b}
	\end{eqnarray}
	where $\mathbf{a},\mathbf{b}\in R^{7\times1}$ and $c^{att}_d \in R^{7\times1}$. 
	
	The maximum value is taken from $c^{att}_d$, and the values for all channels $d=512$ are concatenated to generate a vector $v^{app}$. Channel-wise average values are taken from $f^{tmo}$, and concatenated to generate a vector $v^{tmo}$. We perform the element-wise sum on these two vectors as follows:
	\begin{eqnarray}
	v^{cor}_{d}= \left(max(v_d^{app})+avg(v_{d}^{tmo})\right)
	\end{eqnarray}
	where $v \in R^{512\times1}$ denotes a vector. The element-wise fused features for all channels are concatenated to generate a correlation vector $v^{cor}$. The correlation vector contains correlation cues between the target's action and the camera ego-motion. As a result, the proposed TSCF generates features that are robust against the camera movement by considering the correlations of the three different types of features.

	\subsection{Long Short-Term Memory for Classification}
	To consider the time-series dynamics of videos, we model short-term intervals of videos rather than a single image. To generate feature vectors of the short-term intervals of a video, we first concatenate correlation vectors from times $t$ to $t+L$ as follows:
	\begin{eqnarray}
	{v^{sub,t}}=\left[{v^{cor,t}}, {v^{cor,t+1}},..., {v^{cor,t+L}}\right]
	\end{eqnarray}
	where ${v^{sub,t}}$ is a concatenated vector with $L=3$ times. Then fast fourier transform (FFT) is applied to ${v^{sub,t}}$ as follows:
	\begin{eqnarray}
	v^{fin}=FFT({v^{sub,t}})
	\end{eqnarray}
	The vector $v^{fin} \in R^{2048\times1}$ is used as an input value of the LSTM to track the sequence of the short-term intervals. Then, the LSTM classifies the interaction in the video. As a result, we can calculate the correlation vector encoding on the short-term intervals through the FFT. We can also track the time-series dynamics of the video using the LSTM model.

	\section{Experiments}
	\subsection{Datasets}
	The proposed three-stream fusion network and TSCF were evaluated on two public first-person video datasets: the UTKinect-FirstPerson dataset \cite{utk} and the JPL First-Person Interaction dataset \cite{RYOO2013}. Both datasets are proposed for the purpose of first-person activity recognition. The recognition performances were evaluated on each dataset.
	
	{\bf UTKinect-FirstPerson Dataset (humanoid)}. In this dataset, eight subjects interact with a humanoid robot on which a Kinect sensor is mounted. The human performs friendly, hostile, and normal behaviors toward the robot in a few different background settings. Figure 4 shows an example frame of each of nine classes: hand shake, hug, stand up, wave, point, punch, throw, run, and reach. For the experiments, we used only the RGB frames to allow fair comparisons with the RGB-based competing activity recognition methods. According to the segmented labels, we composed 174 clips from continuous sequences. Half of the clips were chosen as the training data, and the remaining half was used as the testing data. The resolution of the video is $640\times480$.

	\begin{figure}[h]
		\begin{center}
			\includegraphics[width=0.6\linewidth]{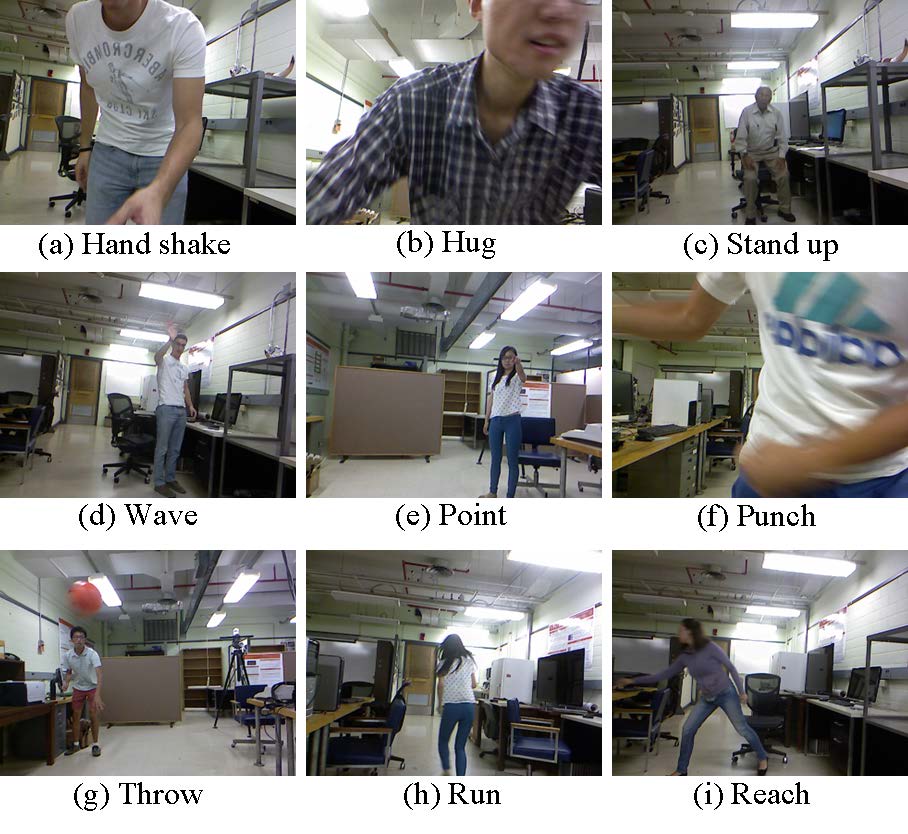}
			\vspace{-0.3cm}
			\caption{\label{framework}Sample frames from the UTKinect-FirstPerson (humanoid) dataset.}
		\end{center}
	\end{figure}

\vspace{-0.4cm}
	
	{\bf JPL First-Person Interaction Dataset}. The dataset was captured using a GoPro2 camera mounted on the head of a humanoid model. The dataset consists of 84 videos with 7 classes: hand shake, hug, pet, wave, point-converse, punch, and throw. Figure 5 shows an example frame of each class. The actions were performed by eight different subjects in various indoor environments and lighting conditions and were recorded at $320\times240$ resolution with 30 fps. The dataset was composed of four positive (hand shake, hug, pet, and wave), one normal (point-converse), and two negative (throw and punch) behavior video clips. For the experiments, we selected half of the videos as the training data and the remaining half as the testing data.
	
	\begin{figure}[h]
		\begin{center}
			\includegraphics[width=0.8\linewidth]{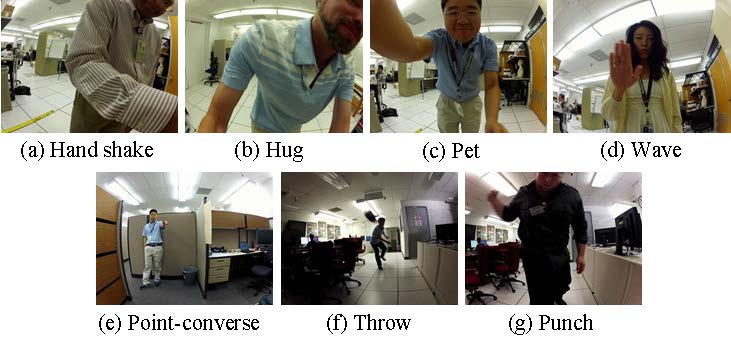}
			\caption{\label{framework}Sample frames from the JPL First-Person Interaction dataset.}
		\end{center}
	\end{figure}

	\subsection{Implementation Details}
	In the three-stream architecture, we utilized the Wang et al.’s VGG-16 \cite{ECCV2016} pre-trained on the UCF101 dataset \cite{ucf}. The pre-trained VGG-16 which contains 13 convolutional layers and 3 fully connected layers was fine-tuned with the UTKinect-FirstPerson dataset and JPL First-Person Interaction dataset such that features were more related to the first-person interaction video. The VGG-16 of the target appearance stream was fine-tuned with the target region of the input images. In this case, the target region of the input image was resized to $224\times224\times3$. The VGG-16 of the target motion stream was fine-tuned with the target region of the optical flow. The non-target region of the optical flow was fed to the ego-motion stream for fine-tuning. The target region and non-target region of the optical flow were resized to $224\times224\times20$. All the images of the training data were used in the training step, and 20 uniformly selected images were used as input images in the testing step. We used the RMSProp optimization algorithm \cite{rms} with a learning rate of $10^{-5}$ and batch size of 20. The VGG-16 of the target appearance stream was trained for 150 iterations. The VGG-16 of each target motion stream and ego-motion stream was trained for 200 iterations. The LSTM model had 700 units in the LSTM cell. The weights and biases of the LSTM were initialized with random values having normal distributions. It was trained for $10^3$ iterations with a 0.9 forget bias and $10^{-4}$ learning rate. We used gradient descent algorithms to optimize the LSTM model. Our CNNs and LSTM were trained to minimize the cross-entropy loss. The implementation of the proposed method was done using python. One NVIDIA GeForce GTX 1080Ti was used to run all the experiments. 
	
	Our final accuracies were the average values for 100 recognition performances. In the experiments, we used the RGB frame and difference of frames separately as an input value for the target appearance stream.
	

	\subsection{Performance Evaluations}
	
	{\bf UTKinect-FirstPerson dataset.} Table 1 shows the recognition performance of the proposed method as compared with that of the competing activity recognition methods on the UTKinect-FirstPerson dataset. The results of long-term recurrent convolutional network (LRCN) \cite{LRCN} and Laptev \textit{et al.} \cite{HOF} are taken from Sudhakaran \textit{et al.} \cite{ICCVW2017}. For the accuracies of a kernelized pooling scheme based on feature subspaces (KRP FS) \cite{che}, we performed experiments on the UTKinect-FirstPerson dataset using the author’s code \cite{url}, except the optical flow. We further compared the proposed method with the Two-Stream ConvNet \cite{TWO}. To implement it, the VGGNet-16 pre-trained on the UCF101 dataset was fine-tuned with the UTKinect dataset.

	\begin{table}[h]
		\label{UTK-LAST_performance}
		\begin{center}
			\begin{tabular}{|l|c|c|c|}
				\hline
				Methods & UTK (\%)\\ 
				\hline\hline
				M. S. Ryoo \textit{et al.} \cite{RYOO2013} & 57.1\\
				I. Laptev \textit{et al.} \cite{HOF}& 48.4\\
				Two-Stream ConvNet \cite{TWO} & 65.9\\
				LRCN (RGB frame) \cite{LRCN} & 72.6\\
				LRCN (difference of frames) \cite{LRCN}& 63.1\\
				KRP FS (RGB frame) \cite{che}& 35.6\\
				KRP FS (difference of frame) \cite{che}& 33.3\\
				S. Sudhakaran \textit{et al.} (RGB frame) \cite{ICCVW2017} & 79.6\\
				S. Sudhakaran \textit{et al.} (difference of frames) \cite{ICCVW2017}& 66.7\\
				\hline
				Ours (RGB frame) & 83.1 \\
				{\bf Ours (difference of frames)} & {\bf84.4}\\
				\hline
			\end{tabular}
		\end{center}
		\caption{Performance comparison of the proposed method with competing methods on the UTKinect-FirstPerson dataset.} 
	\end{table}

	
	In the UTKinect-FirstPerson dataset, the video clips contain numerous camera ego-motions that are not related to the target’s action. Therefore, the difference of frames can play the role of noise to analyze the target’s action in the UTKinect-FirstPerson dataset. In Table 1, the competing methods show a higher performance when the RGB frame instead of the difference of frames is used as the input value. However, our proposed method outperformed all the competing methods, whether the input value was the RGB frame or the difference of frames. This is because the TSCF of the proposed network generates robust features that mitigate the effects of camera ego-motion. In Figure 6, classes such as hand shake, hug, and punch, where a relatively large amount of camera ego-motion occurs, show a higher performance. These results show that our proposed method can recognize interactions in first-person video, although a large amount of camera ego-motion occurs. Figure 6 also shows that the recognition of the point and reach classes remain difficult. The point, wave, reach, and throw classes are similar, because the target is standing and stretching his arm. These actions should be more elaborately considered, reflecting the shape of the target's hand or the motion of the arm.

	
	As compared to the previous competing first-person interaction recognition methods  \cite{RYOO2013, ICCVW2017}, our method shows significantly improved accuracy on the UTKinect-FirstPerson dataset. Furthermore, we compared three recent activity recognition methods \cite{TWO,LRCN, che} that do not consider the camera ego-motion, because they were developed based on third-person video. The results show that the proposed method outperformed the competing methods by considering the camera ego-motion for the first-person interaction recognition. 
	
		\begin{figure}[t]
		\begin{center}
			\includegraphics[width=0.8\linewidth]{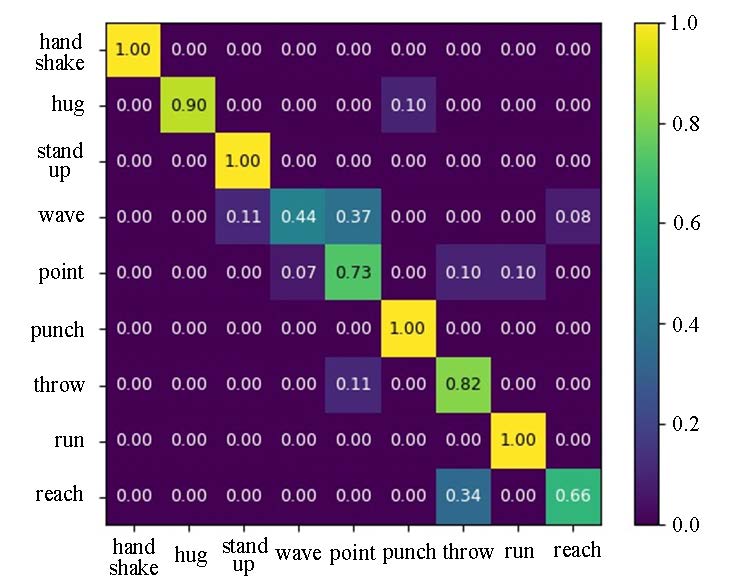}
			\caption{\label{utk-confusion} Confusion matrix of the proposed three-stream fusion network on the UTKinect-FirstPerson dataset.}
		\end{center}
	\end{figure}

	{\bf JPL First-Person Interaction dataset.} The recognition accuracies of the proposed method and competing activity recognition methods on the JPL First-Person Interaction dataset are shown in Table 2. The LRCN results \cite{LRCN} are taken from Sudhakaran \textit{et al.} \cite{ICCVW2017}. The results of KRP FS \cite{che} are reproduced by using the author's code \cite{url}. To obtain the results of two-stream ConvNet \cite{TWO}, we fine-tuned the VGGNet-16 which was pre-trained on the UCF101 dataset.
	
	\begin{table}[h]
	\label{JPL-LAST_performance}
	\begin{center}
		\begin{tabular}{|l|c|c|c|}
			\hline
			Methods & JPL (\%)\\
			\hline\hline
			M. S. Ryoo \textit{et al.} \cite{RYOO2013} & 89.6 \\
			Boosted MKL \cite{KERNEL}& 87.4 \\
			Two-Stream ConvNet \cite{TWO} & 54.2\\
			LRCN (RGB frame) \cite{LRCN} & 59.5 \\
			LRCN (difference of frames) \cite{LRCN} & 89.0 \\
			KRP FS (RGB frame) \cite{che}& 73.8 \\
			KRP FS (difference of frames) \cite{che}& 85.7 \\
			SeDyn \cite{SUB-EVENT} + FTP \cite{ftp} & 92.9 \\
			S. Sudhakaran \textit{et al.} (RGB frame) \cite{ICCVW2017} & 70.6 \\
			S. Sudhakaran \textit{et al.} (difference of frames) \cite{ICCVW2017}& 91.0 \\
			\hline	
			Ours (RGB frame) & 88.0 \\
			{\bf Ours (difference of frames)} & {\bf94.4} \\
			\hline
		\end{tabular}
	\end{center}

	\caption{Performance comparison of proposed method with competing methods on the JPL First-Person Interaction dataset.} 
\end{table}

	As shown in Table 2, Sudhakaran \textit{et al.} \cite{ICCVW2017}, LRCN \cite{LRCN}, KRP FS \cite{che}, and our proposed method showed a higher performance when the difference of frames was used instead of the RGB frame as the input value. We note that the recognition performance with the difference of frames is much higher than with the RGB frame on the JPL First-Person Interaction dataset than on the UTKinect-FirstPerson dataset. This is because the JPL First-Person Interaction dataset consists of camera ego-motion related to the target’s actions. The videos were captured while the camera was fixed to the wearer, and the camera wearer rarely moved by himself. Owing to the relatively stable camera movement, we believe that the difference of frames is a better representation of the motion pattern for two consecutive frames. As a result, the proposed method with the difference of frames outperformed other competing methods in the experiments. In addition, Figure 7 illustrates that our proposed method performed well for all action classes. These results show that our method is effective regardless of the existence of camera movement. However, some wave class videos, where the target waves his hand close to the camera wearer, were recognized as the pet class, as seen in Figure 7. We assume that these results occurred because a large number of motion vectors caused by a waving hand appeared in the target regions of the optical flow as in the pet class.

	\begin{figure}[t]
		\begin{center}
			\includegraphics[width=0.8\linewidth]{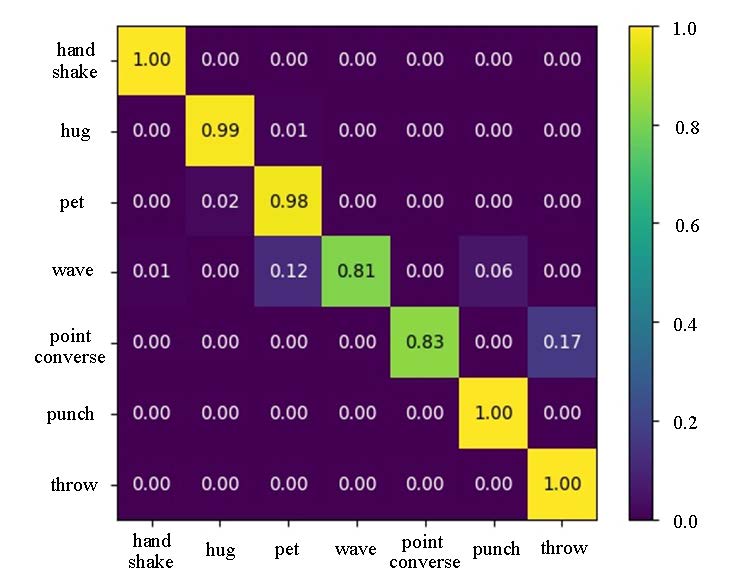}
			\caption{\label{JPL-confusion} Confusion matrix of the proposed three-stream fusion network on the JPL First-Person Interaction dataset.}
		\end{center}
	\end{figure}

	\subsection{Ego-Motion Stream Evaluations}
	To validate the contribution of using camera ego-motion, we compared the proposed method and the two-stream (target appearance stream and target motion stream) method, excluding only the ego-motion stream in the implementation of the proposed method. The two-stream method was fused by using Eqs. (1), (2), and (3) in order. In Table 3, it can be seen that the proposed method that considers the camera ego-motion obtains the best performance. However, the two-stream (difference of frames) method on the UTKinect dataset obtains a 14.9\% accuracy performance. In first-person video, the appearance of the target is easily distorted when the camera shakes a considerable amount. Because the camera shake is greater in the UTKinect dataset than in the JPL dataset, the difference of frames on the UTKinect dataset can act as noise. These results show that it is difficult to recognize interactions using only target appearance and target motion features. In addition, the effect of the ego-motion stream is more pronounced in the UTKinect dataset, where the camera shakes considerably. The results in Table 3 show that it is important to consider the camera ego-motion in first-person interaction recognition.
	
		\begin{table}[h]
		\label{UTK-LAST_performance}
		\begin{center}
			\renewcommand{\tabcolsep}{1mm}
			\begin{tabular}{|l|c|c|c|}
				\hline
				Stream & UTK & JPL\\ 
				\hline
				Two-stream (RGB frames) &67.1&68.9\\
				Two-stream (Difference of frames) &14.9&74.0\\
				\hline
				Three-stream fusion network (RGB frame) & 83.1&88.0\\
				{\bf Three-stream fusion network (Difference of frames)}& {\bf84.4}&{\bf94.4}\\
				\hline
			\end{tabular}
		\end{center}
		\caption{Accuracies on the two-stream and the three-stream fusion networks.} 
	\end{table}

	\subsection{Three-Stream Correlation Fusion Evaluations}
	To validate the proposed fusion method called TSCF, we compared the TSCF and frequently used fusion methods. Table 4 shows the performances when the three-stream fusion network used common fusion methods instead of the TSCF. In Eq. (4) $v^{cor}_{d}$, we used the channel-wise maximum values of the feature maps combined by the common fusion method. The table shows that the TSCF is the most suitable method for fusing the features of three-stream architecture.
	

	\begin{table}[h]
		\label{fusion}
		\begin{center}
			\begin{tabular}{|l|c|c|c|}
				\hline
				Fusion methods & UTK & JPL\\ 
				\hline
				Sum \cite{FUSION} (RGB frames)&81.7&86.3\\
				Sum \cite{FUSION} (Difference of frames)&83.7&92.5\\
				Max \cite{FUSION} (RGB frames)&82.4&87.3\\
				Max \cite{FUSION} (Difference of frames)&81.8 &85.8\\
				Bilinear \cite{lin2015bilinear} (RGB frames)&74.4&77.8\\
				Bilinear \cite{lin2015bilinear} (Difference of frames)&74.7&77.6\\
				\hline
				TSCF(RGB frames)&83.1&88.0\\
				{\bf TSCF(Difference of frames)}&{\bf84.4}&{\bf94.4}\\
				\hline
			\end{tabular}
		\end{center}
		\caption{Accuracies on various fusion methods and the three-stream correlation fusion.}
	\end{table}

	
	
	Further, we demonstrated the effectiveness of the proposed TSCF for the first-person interaction recognition as compared to other features. The t-SNE technique \cite{tsne} was used to visualize the feature vectors of the target appearance stream, the target motion stream, the ego-motion stream, and our TSCF method. To obtain the results for our TSCF, the t-SNE was applied to the input value of the LSTM. For the three streams, channel-wise average values were taken from each feature map and concatenated as a vector. Then, we applied the aforementioned Eqs. (5) and (6) to the concatenated vector, and obtained the t-SNE results.
	

	
		\begin{figure}[]
		\begin{center}
			\includegraphics[width=1.0\linewidth]{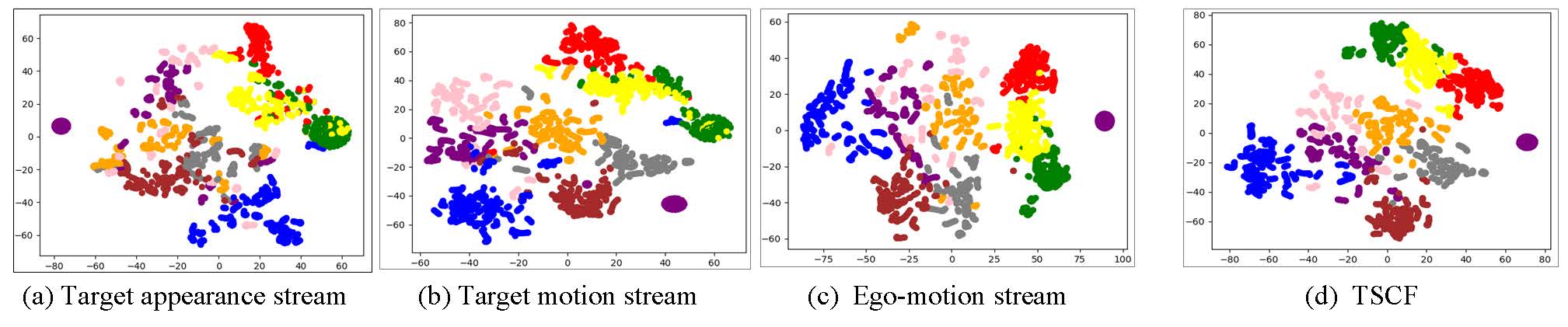} 
			\vspace{-1.0cm}
			\caption{\label{utk-tsne}t-SNE results for the UTKinect-FirstPerson (humanoid) dataset. (a), (b), and (c) represent the feature vectors of the target appearance stream, the target motion stream, and the ego-motion stream, respectively. (d) represents the feature vectors of the three-stream correlation fusion. The color list for the nine classes is as follows: hand shake(red), hug(green), stand up(blue), wave(pink), point(purple), punch(yellow), throw(orange), run(brown), and reach(gray).}
		\end{center}
	\end{figure}
	
	\begin{table}[]
		\label{UTK_performance}
		\begin{center}
			\begin{tabular}{|l|c|c|c|}
				\hline
				Stream of the three-stream architecture & UTK (\%)\\ 
				\hline\hline
				Target appearance stream & 60.5\\
				Target motion stream & 78.3\\
				Ego-motion stream & 75.6 \\
				\hline
				TSCF & 84.4 \\
				\hline
			\end{tabular}
		\end{center}
		\vspace{-0.4cm}
		\caption{Performance comparison of each target appearance stream, target motion stream, ego-motion stream, and three-stream correlation fusion on the UTKinect-FirstPerson dataset.} 
	\end{table}

	Figure 8 shows the t-SNE results on the UTKinect-FirstPerson dataset. As compared to Figure 8(a), Figure 8(d) shows that the feature vectors of nine classes are less discriminative. Compared to Figure 8(b), Figure 8(d) is a better representation for the hug and punch classes, where a large amount of camera ego-motion occurs, because the target directly interacts with the camera wearer at a close distance. As compared to Figure 8(c), Figure 8(d) has more discriminative feature vectors for the stand up, wave, point, and throw classes, where some camera ego-motion occurs, because the target acts at a long distance from the camera wearer. In Table 5, we additionally compare the performance of our fusion method with individual single streams of the three-stream architecture. Our fusion method shows a higher performance than the other features.

\begin{figure}[]
	\begin{center}
		\includegraphics[width=1.0\linewidth]{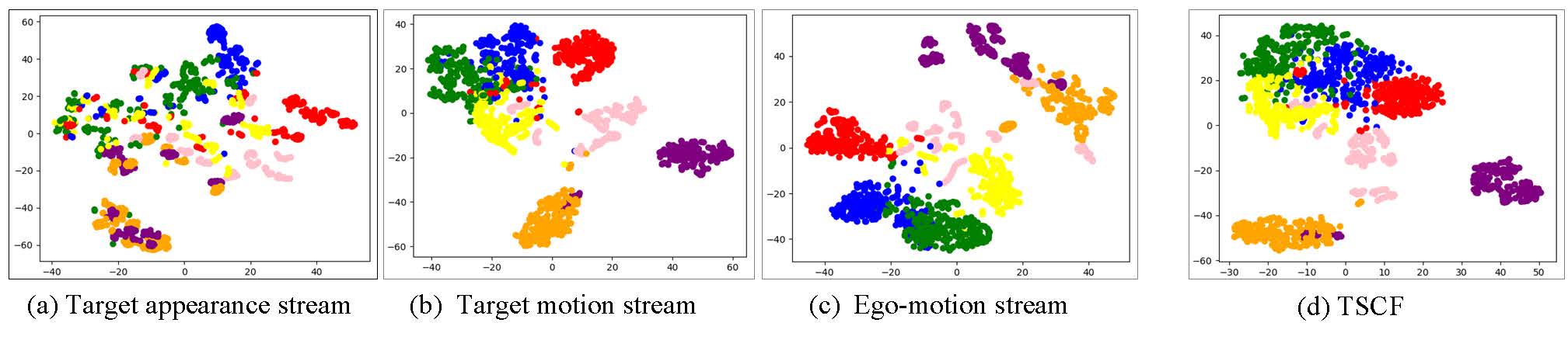} 
		\vspace{-1.0cm}
		\caption{\label{jpl-tsne}t-SNE results for the JPL First-Person Interaction dataset. The results in (a), (b), and (c) represent the feature vectors of the target appearance stream, target motion stream, and ego-motion stream, resepectively. (d) represents the feature vectors of the three-stream correlation fusion. The color list for seven classes is as follows: hand shake(red), hug(green), pet(blue), wave(pink), point-converse(purple), punch(yellow), and throw(orange).}
	\end{center}
\end{figure}

\begin{table}[]
	\label{JPL_performance}
	\begin{center}
		\begin{tabular}{|l|c|c|c|}
			\hline
			Stream of the three-stream architecture & JPL (\%)\\ 
			\hline\hline
			Target appearance stream & 68.7 \\
			Target motion stream & 92.8 \\
			Ego-motion stream & 88.1 \\
			\hline
			TSCF & 94.4 \\
			\hline
		\end{tabular}
	\end{center}
	\vspace{-0.4cm}
	\caption{Performance comparison for each target appearance stream, target motion stream, ego-motion stream, and the proposed fusion method on the JPL First-Person Interaction dataset.}
\end{table}

	We also evaluated our TSCF on the JPL First-Person Interaction dataset, the results of which are shown in Figure 9. The feature vectors of Figure 9(a) are less discriminative for all seven classes than those in Figure 9(d). Compared to Figure 9(b), Figure 9(d) has more discriminative feature vectors for the hug and pet classes where the target appearance is similar and a large amount of camera ego-motion occurs. As compared to Figure 9(c), Figure 9(d) is a better representation for the wave, point-converse, and throw classes, where little camera ego-motion occurs. In Table 6, it can be seen that our fusion method shows a higher performance than a method using any one of the three streams alone.

	In all of the aforementioned datasets, our TSCF obtained the highest performance. Interestingly, the results of the ego-motion stream show that the camera ego-motion can be an important clue for analyzing first-person interaction. These results show that consideration of the correlations of the different three types of features can improve the discriminative power of the feature vectors. Although we fuse three different types of features, our TSCF rearranges these three feature types and generates more discriminative features to handle first-person interaction video. In other words, our TSCF can compensate for target appearance, target motion, and camera ego-motion features by considering the correlations of these three types of features, which have their own characteristics.

	\subsection{Correlation Evaluations}
	The proposed TSCF considers the correlation between a human's appearance and motion by generating the sum of the appearance and motion feature maps. It also considers the correlation between the target's motion and the camera wearer's motion by summing the motion and ego-motion feature maps. Finally, we consider the correlation between the target's behavior and the camera wearer's movement by generating  the correlation vector. To validate that it is important to consider these correlations for first-person interaction recognition, we compared the proposed TSCF and the three-stream deep features \cite{kim2018first}. After the channel-wise maximum value and the channel-wise average value are obtained from the feature maps, the three-stream deep features technique \cite{kim2018first} sums all these values to make features for first-person interaction recognition.
	
	Table 7. shows the recognition accuracy of the three-stream deep features \cite{kim2018first} on the UTKinect-FirstPerson dataset and the JPL First-Person Interaction dataset. We performed experiments to obtain the results of the three-stream deep features \cite{kim2018first} on the JPL First-Person Interaction dataset. As shown in Table 7, the three-stream deep features \cite{kim2018first} obtains similar accuracies when the raw frame and the difference of frames are used by considering the features of target appearance, target motion, and camera ego-motion. However, three-stream deep features shows lower accuracies than the proposed method. The results show that it is important to consider the correlations in first-person interaction recognition.

		\begin{table}[h]
		\label{fusion}
		\begin{center}
			\begin{tabular}{|l|c|c|c|}
				\hline
				Fusion methods & UTK & JPL\\ 
				\hline
				Three-stream deep fusion \cite{kim2018first} (RGB frames)&80.37&81.1\\
				Three-stream deep fusion \cite{kim2018first} (Difference of frames)&86.5&86.4\\
				\hline
				TSCF(RGB frames)&83.1&88.0\\
				{\bf TSCF(Difference of frames)}&{\bf84.4}&{\bf94.4}\\
				\hline
			\end{tabular}
		\end{center}
		\vspace{-0.4cm}
		\caption{Accuracies of the three-stream deep fusion and the three-stream correlation fusion.}
	\end{table}

	

	\section{Conclusion}
	We proposed a three-stream fusion network for interaction recognition in first-person videos where camera ego-motion occurs. The TSCF was introduced to consider the correlations of target appearance, target motion, and camera ego-motion features. The proposed three-stream fusion network with the TSCF successfully classified the first-person interaction video clips by means of robust video feature vectors that mitigate the effects of the camera’s movement. The proposed method showed a state-of-the-art performance on the UTKinect-FirstPerson dataset and the JPL First-Person Interaction dataset. Performance comparisons with third-person video-based approaches showed that consideration of the camera ego-motion is an important aspect of first-person interaction recognition. In addition, comparisons with other fusion methods showed the effectiveness of the proposed TSCF which considers the correlations between target appearance, target motion, and camera ego-motion. Furthermore, the three-stream fusion network can recognize first-person interactions when the the camera movement occurs by showing a higher accuracy when the difference of frames is used as the input value. In the future, we will focus on the recognition of interaction classes where the target performs certain actions, such as \textit{point} and \textit{wave}, at a distance from the camera wearer.

	\section*{Acknowledgment}
	This work was supported by Institute for Information \& Communcations Technology Planning \& Evaluation(IITP) grant funded by the Korea goverment(MSIT) [No. 2019-0-00079, Department of Artificial Intelligence, Korea University] and [No. 2014-0-00059, Development of Predictive Visual Intelligence Technology].
	
	
	\bibliography{egbib_3}
	
\end{document}